\documentclass[journal,comsoc]{IEEEtran}
\usepackage{booktabs}
\usepackage{graphicx} 
\include{diagrams}
\usepackage{float}
\usepackage{array} 
\usepackage[thinlines]{easytable}
\usepackage{amsmath}
\usepackage{multirow}
\usepackage[caption=false]{subfig}
\usepackage{mathtools}
\usepackage{amssymb}
\usepackage{listings}
\usepackage{wasysym}
\usepackage[table,xcdraw]{xcolor}
\usepackage{varwidth}
\usepackage{hyperref}
\usepackage{footnote}
\usepackage{threeparttable}
\usepackage{makecell}
\usepackage{soul}
\usepackage{algorithm}
\usepackage{algorithmic}

\DeclareCaptionFormat{mycaption}{%
  \begin{varwidth}{\linewidth}%
    \centering
    #1#2#3%
  \end{varwidth}%
}

\newcolumntype{L}[1]{>{\raggedright\let\newline\\\arraybackslash\hspace{0pt}}m{#1}}
\newcolumntype{C}[1]{>{\centering\let\newline\\\arraybackslash\hspace{0pt}}m{#1}}
\newcolumntype{R}[1]{>{\raggedleft\let\newline\\\arraybackslash\hspace{0pt}}m{#1}}

\makeatletter
\newlength{\@captionlabelwidth}
\DeclareRobustCommand*{\centeredmultilineincaption}[1]{%
  \settowidth{\linewidth}{%
    \@nameuse{fnum@\@captype}: %
  }%
  \begin{tabular}[t]{@{\hspace{-\@captionlabelwidth}}c@{}}%
    \hspace{\@captionlabelwidth}\ignorespaces
    #1%
  \end{tabular}%
}

\ifCLASSOPTIONcompsoc
  \usepackage[nocompress]{cite}
\else
  \usepackage{cite}
\fi

\ifCLASSINFOpdf
\else
\fi
\begin{document}
%
\title{A physics-informed and attention-based graph learning approach for regional electric vehicle charging demand prediction}
%
%

\author{Haohao Qu, Haoxuan Kuang, Jun Li, Linlin You$^*$,~\IEEEmembership{Member,~IEEE,}\\
\IEEEcompsocitemizethanks{\IEEEcompsocthanksitem H. Qu, H. Kuang, J. Li, and L. You are with the School of Intelligent Systems Engineering, Sun Yat-Sen University, Guangzhou, 510006, CN. H. Qu is also wih the Department of Computing, The Hong Kong Polytechnic University, Hong Kong, 100872, CN.
\protect\\
}
\thanks{*Corresponding author, e-mail: youllin@mail.sysu.edu.cn}
}

%
%

\markboth{IEEE TRANSACTIONS ON INTELLIGENT TRANSPORTATION SYSTEMS,~Vol.~XX, No.~X, XX~2021}%
{Qu \MakeLowercase{\textit{et al.}}: A physics-informed and attention-based graph learning approach for regional electric vehicle charging demand prediction}
%



\maketitle

\begin{abstract}
Along with the proliferation of electric vehicles (EVs), optimizing the use of EV charging space can significantly alleviate the growing load on intelligent transportation systems. As the foundation to achieve such an optimization, a spatiotemporal method for EV charging demand prediction in urban areas is required. Although several solutions have been proposed by using data-driven deep learning methods, it can be found that these performance-oriented methods may suffer from misinterpretations to correctly handle the reverse relationship between charging demands and prices. To tackle the emerging challenges of training an accurate and interpretable prediction model, this paper proposes a novel approach that enables the integration of graph and temporal attention mechanisms for feature extraction and the usage of physic-informed meta-learning in the model pre-training step for knowledge transfer. Evaluation results on a dataset of 18,061 EV charging piles in Shenzhen, China, show that the proposed approach, named PAG, can achieve state-of-the-art forecasting performance and the ability in understanding the adaptive changes in charging demands caused by price fluctuations.

\end{abstract}


%
\IEEEpeerreviewmaketitle

\section{Introduction}\label{sec:intro}

\IEEEPARstart{D}RIVEN by the public concern about climate change and the support of government policy, the global sales of electric vehicles (EVs) have kept rising strongly over the past few years \cite{Charging2022Powell}. According to the International Energy Agency (IEA), global public spending on EV subsidies and incentives doubled to nearly USD 30 billion in 2021 \cite{GlobalEVOutlook2022}. Such rapid growth not only resulted in a net reduction of 40 million tons of carbon dioxide but also led to increased loads on urban transportation systems \cite{Hidden2023Ren}. In this context, various EV charging-related smart services are developed and applied to facilitate the development of Intelligent Transportation Systems (ITS), e.g., with accurate regional EV charging demand prediction, proper parking recommendations can be provided to en-route drivers to mitigate mileage anxiety \cite{ChargingRec2016Tian}, and dynamic pricing schemes can be applied by regulators to improve energy efficiency \cite{Dynamic2021Fang}.

\begin{figure}
	\centering
	\includegraphics[width=3.5in]{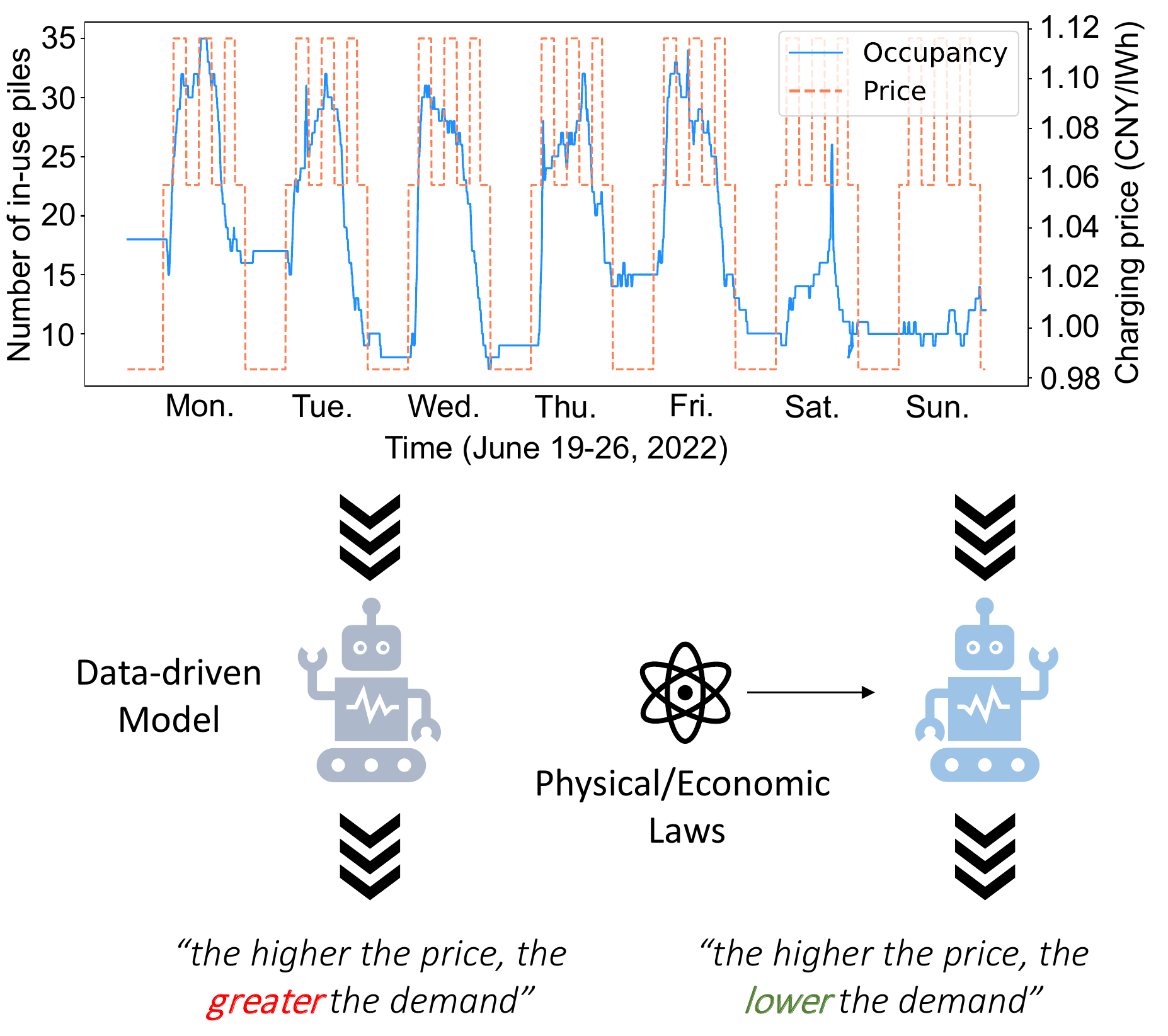}
	\caption{Misinterpretation of EV charging demand caused by price fluctuations.}
	\label{fig:misinterpretation}
\end{figure}

With the development of urban road networks, spatiotemporal EV charging demand prediction is of growing interest to accommodate the increasing connectivity among urban areas (i.e., zones). Such that, two challenges are emerging. First, compared to conventional methods, feature extraction capabilities shall be enhanced to capture not only temporal patterns hidden in the time series data but also spillover effects propagating among neighboring areas. More importantly, existing data-driven models, which mainly focus on reducing forecasting errors, suffer from misinterpretations as illustrated in Fig. \ref{fig:misinterpretation}, where a higher price may be interpreted by the model to have a higher demand, as the partially observed data without the knowledge presented by the common physical/economic laws are used. Even though such data-driven models can still perform well in most cases, the misinterpretation has fatal flaws to support decision-making processes, which shall be causally reflecting the fact that ''a higher price will impact the overall demand''.

In order to make full use of spatial and temporal data, various studies are proposed to design lower error models by integrating graph convolutional networks (GCNs) with recurrent neural networks (RNNs) \cite{STCGCN2023Bao,Modeling2023Irfan}, but they are criticized for the inflexibility and ambiguity in weight assignment due to the limitation within convolutional and recurrent operations. Recently, a more advanced technique, called Attention Mechanism, \cite{Attention2017Ashish}, is widely discussed and utilized to enhance such integration with a stronger feature representation capability \cite{TPA2019Shih,GAT2018Petar}. However, the corresponding network can still not cover the dependencies and interactions between EV charging-related factors. As for misinterpretations, Physics-informed Neural Networks (PINNs) \cite{PINN2019Raissi} and Meta-learning \cite{Meta2017Finn} are two feasible solutions. Specifically, PINNs can encode explicit laws of physics, while meta-learning can facilitate the transformation of knowledge learned from related domains. However, the approach with PINN, Meta-learning, GCN, and attention mechanism integrated is still missing to support accurate and unbiased knowledge learning.

To fill the gap, this paper proposes a physics-informed and attention-based graph learning approach for the prediction of EV charging demand, called PAG, which combines Physics-Informed Meta-Learning (PIML), Graph Attentional Network (GAT), and Temporal Pattern Attention (TPA) for model pre-training, graph embedding, and multivariate temporal decoding, respectively. Specifically, on the one hand, the typical convolutional and recursive operations are replaced by the two attention mechanisms to enhance the flexibility and interpretability of the prediction model. On the other hand, the pre-training step (i.e., PIML) processes tuning samples generated according to predefined physic/ economic laws and the observed samples together to extract common knowledge through meta-learning with potential misinterpretations remedied.

Moreover, through a rigorous evaluation based on a real-world dataset with samples of 18,061 EV charging piles in Shenzhen, China, from 19 June to 18 July 2022 (30 days), the efficiency and effectiveness of the proposed model are demonstrated. Specifically, as shown by the results, PAG can reduce forecasting errors by about 6.57\% in four metrics, and compared to other state-of-the-art baselines, it can also provide correct responses to price fluctuations.

In summary, the main contributions of this paper include:

\begin{enumerate}
	\item An effective prediction model combing two attention mechanisms, i.e., GAT and TPA, is designed for EV charging demand prediction.  
	
	\item A informed model pre-training method, called PIML, is proposed to tackle the intrinsic misinterpretations of conventional data-driven methods to mine the correct relationship between the charging demands and prices.  
	
	\item An integration of GAT, TPA, and PIML enables knowledge adaptation and propagation to build an outstanding predictor with high accuracy and correct interpretation.
\end{enumerate}

The remainder of this paper is structured as follows. Section \ref{sec:review} summarizes related challenges and solutions about EV charging demand prediction. Then, Section \ref{sec:method} introduces the proposed approach PAG, which is evaluated in Section \ref{sec:evaluation}. Finally, conclusions and future works are drawn in Section \ref{sec:conclusion}.

\section{Literature review}\label{sec:review}
In this section, related challenges together with solutions for EV charging demand prediction are summarized.

\subsection{Emerging challenges} \label{subsec:challenges}
In general, to create an efficient and effective model with temporospatial information embedded for EV charging demand prediction and analysis, several challenges are emerging, which include:

\begin{itemize}	
	\item Feature extraction: EV charging stations are scattered all over the city and correlated with each other, as the temporal demands can only be fulfilled once and the price and location differences may influence their utilization. By considering charging stations as nodes, their relationships (such as distances) as edges, time-varying demand as sequences, and influencing factors as features, how to capture the hidden patterns in such a high-dimensional tensor with heterogeneous components is the key issue in the EV charging demand prediction.
	
	\item Biased information: The data-driven models for EV charging demand prediction can suffer from an intrinsic problem of misinterpretation, i.e., "the higher the price the higher the occupancy" as illustrated by Fig. \ref{fig:misinterpretation}, which is caused by the partially observed data associated with mass-deployed pricing strategies of reducing peak charging demands by raising prices. In the case that unbiased data is hard to be obtained, how to counterbalance the influence of biased information is an important problem to ensure the correctness of forecasting results.
	
	\item Knowledge learning: The relationship between EV charging demand and related factors vary across time and place. When introducing different physical and economic laws to equip data-driven models \cite{PIST2023Shi}, a unified and scalable method is required to learn the knowledge fairly from observed and tuning samples (generated from physical and economic laws).
\end{itemize}

\subsection{Related solutions}
To tackle the aforementioned challenges, several solutions are proposed. Initially, to address graph embedding and multivariate time-series forecasting, scholars mainly focus on statistic reasoning of model graph structures and infer future statuses, separately \cite{Rapid2012Zhou,STPower2012Luk}. For instance, a spatiotemporal model is proposed \cite{STEV2014Mu} to evaluate the impact of large-scale deployment of plug-in electric vehicles (EVs) on urban distribution networks. Although these statistical reasoning-based models can be computationally efficient and easy to be interpreted, there is still a limitation in expressing high-dimensional and non-linear features.

More recently, with the rapid development of deep learning techniques, the integration of Graph Neural Networks (GNNs) and Recurrent Neural Networks (RNNs) is emphasized to enable accurate spatiotemporal prediction in transportation systems, such as parking occupancy \cite{Parking2019Yang}, traffic speed \cite{STTGCN2023Xu}, flow \cite{STMeta2022Zhang} and also EV charging demand predictions \cite{STSimulation2019Xiang,STdistribution2020Yi}, e.g., SGCN \cite{Operating2023Su} combines graph convolutional network (GCN) and gated recurrent unit (GRU) to extract spatial and temporal features of the operating status at EV charging stations, respectively to better assist the prediction. However, the typical convolutional and recurrent structures can be crippled by their inflexibility and ambiguity in weight assignment.

Later on, a more advanced technique, known as attention mechanism \cite{Attention2017Ashish}, and its variants \cite{TPA2019Shih,GAT2018Petar} pose the potential in refining the typical network structures, due to their strong feature representation capabilities. Solutions in spatiotemporal prediction include AST-GAT \cite{ASTGAT2022Li} and STANN \cite{STANN2019Loan}. Both of them use spatial and temporal attention mechanisms to assign appropriate weights to the dependencies between road segments and time steps. Nevertheless, a similar integration of spatiotemporal attention is still missing to capture the interaction of EV charging-related factors.

More importantly, although these performance-oriented deep learning methods have made significant contributions to the prediction tasks in intelligent transportation systems, they still have key weaknesses in addressing misinterpretations while full samples can not be sensed. To address such a problem, Physic-Informed Neural Networks (PINNs) \cite{PINN2019Raissi} are proposed and discussed, which can regulate model training processes by introducing an extra loss based on given physical laws, e.g., an accurate model for urban traffic state estimation can be trained with a hybrid loss function that can measure not only the spatiotemporal dependence but also the internal law of traffic flow propagation \cite{PIST2023Shi}. However, knowledge learning within PINNs remains challenging, especially when the laws can not be explicitly expressed. 

As a potential solution, meta-learning \cite{Meta2017Finn} starts to be utilized, which can transfer knowledge from other domains and migrate gradients to update the global model, making the learning process smoother\cite{Transfer2010Pan}. Taking a study on urban traffic dynamics prediction as an example \cite{STMeta2022Zhang}, it shows that even a simple integration of Convolutional Neural Network (CNN) and Long Short-Term Memory (LSTM) can achieve state-of-the-art performance after being pre-trained by meta-learning. This advanced technique provides inspiration to address misinterpretations in EV charging demand prediction, but its limitation is the need to acquire representative data for pre-training, which is expensive and laborious. 


In summary, even though the combination of GNNs and RNNs makes significant achievements in prediction tasks to assist intelligent transportation systems, the introduction of attention-based mechanisms provides additional capabilities in improving feature representation. Moreover, PINNs and meta-learning can enable knowledge transfer to remedy the impact of misinterpretation, however, how to balance the uncertainty and the learning cost is still challenging. Therefore, this paper proposes a novel approach, called PAG, which integrates GAT, TPA, and PIML to build an accurate and robust model for EV charging demand prediction.

\section{Proposed approach} \label{sec:method}
\begin{figure*}
	\centering
	\includegraphics[width=\textwidth]{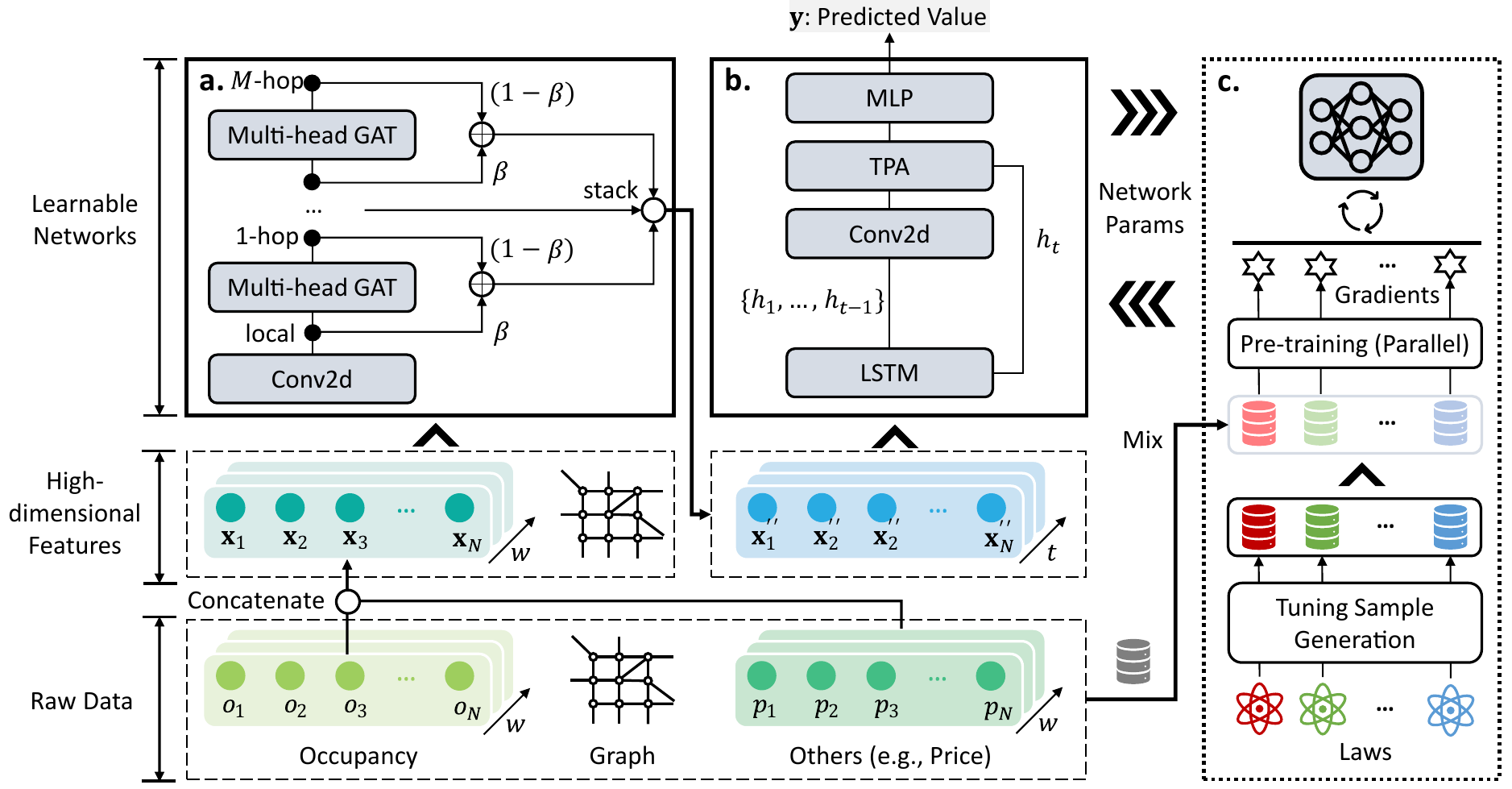}
	\caption{Overall structure of the proposed approach, which consists of \textbf{(a)}. Graph Embedding Module, \textbf{(b)}. Multivariate Decoder Module, and \textbf{(c)}. Model Pre-training Module.}
	\label{fig:overall}
\end{figure*}

As illustrated in Fig. \ref{fig:overall}, the proposed approach consists of three modules, namely \textbf{a)} a graph embedding module, which transfers spatial relationships between nodes to dense vectors based on GAT; \textbf{b)} a multivariate decoder module, which processes hop-wise dense vectors by a neural network of TPA-LSTM for the predicted value; and \textbf{c)} a model pre-training module, which optimizes the network parameters to avoid misinterpretations by applying a law-based pseudo-sampling and meta-learning strategy. In the following sections, the prediction problem will be first defined, and then, the details of each module will be described.

\subsection{Problem definition}
For consistency, the paper uses the lowercase letters (e.g., $x$) to represent scalars, bold lowercase/Greek letters with the upper arrow (e.g., $\mathbf{x}$ and $\vec{\alpha}$) to denote column vectors, bold-face upper case letters (e.g., $\mathbf{X}$) to denote matrices and uppercase calligraphic letters (e.g., $\mathcal{X}$) to denote sets or high-order tensors.

Moreover, given a matrix $\mathbf{X}$, $\mathbf{X}(i,:)$ and $\mathbf{X}(:,j)$ represent its $i_{th}$ row and $j_{th}$ column, respectively, and the ($i_{th}, j_{th}$) entry of $\mathbf{X}$ can be denoted as $\mathbf{X}(i,j)$. $\mathbf{X}^{\rm T}$ and $\mathbf{x}^{\rm T}$ are used to represent the transposes of $\mathbf{X}$ and $\mathbf{x}$, respectively. Besides, $\mathbb{R}$ is used to represent the data space, e.g., $\mathbf{X} \in \mathbb{R}^{M \times N}$ means that $\mathbf{X}$ has two dimensions of $M$ and $N$. The element-wise product of $\mathbf{x}$ and $\mathbf{y}$ is represented as $\mathbf{x} \otimes \mathbf{y}$, and their concatenation is noted as $\mathbf{x} \Vert \mathbf{y}$. 

In this paper, the EV charging demands are distributed spatiotemporally in the urban area, and the demands are not only directly related to past charging demands but also indirectly influenced by other demand-sensitive features, e.g., charging prices. Therefore, the problem can be defined as a multivariate spatiotemporal forecasting problem with the following settings:

\begin{enumerate}
	\item Data structure: The studied urban area can be structured as a graph, which has $N$ zones as its nodes and the links of their centroids as edges. Accordingly, a three-dimension tensor $\mathcal{X}$ is used to manage the data, e.g., $\mathcal{X}(i,j,t)$ represents $j_{th}$ feature for $i_{th}$ node at time $t$;
	
	\item Time-series input: To support the training of data-driven models, the sliced data with a window size $w$ are used, which is denoted as $\{ [\mathbf{x}_{t-w}, \mathbf{x}_{t-w+1}, ... \mathbf{x}_{t-1} \}$;
	
	\item Task objective: It is to predict the future distribution of EV charging demand, i.e., $\mathbf{y}=\mathcal{X}(:,o,t+\Delta)=\mathbf{o}_{t+\Delta}$, where $\Delta$ is a fixed time interval for forecasting, and $\mathbf{o}_{t+\Delta}$ represents charging demands for all studied traffic zones at time $(t+\Delta)$.
\end{enumerate}

\subsection{Graph Embedding Module}
To better extract the internal and external relationships among node features, the graph embedding module is designed with two networks, i.e., CNN and GAT. First, CNN is used to extract the temporal features by employing a 2-dimensional convolution kernel \cite{CNN1989LeCun}, whose width is equal to the number of features in each node. Note that its height and stride are two hyperparameters (by default 2 and 1, respectively). Second, to model the spatial relationships among studied zones (i.e., districts), GAT with a masked multi-head attention mechanism is used to strengthen the extraction of neighbor features. 

As for the propagation process of each GAT layer, the module processes the node features (i.e., $\mathbf{X}=\{\mathbf{x}_1, \mathbf{x}_2, ..., \mathbf{x}_N\}, \mathbf{x}_i \in \mathbb{R}^F$, $N$ is the number of nodes, and $F$ is the feature number of each node) for the output (i.e., $\mathbf{X}'=\{\mathbf{x}_1', \mathbf{x}_2', ..., \mathbf{x}_N'\}, \mathbf{x}_i' \in \mathbb{R}^{F'}$) in three steps, namely:

\textbf{Step 1:} For node $i$, its similarity with each neighbor $j \in \mathcal{N}_i$ (neighors for node $i$ in the graph) is calculated according to Formula \eqref{eq:similarity}. 

\begin{equation}
	\label{eq:similarity}
	\mathbf{e}_{ij} = a([\mathbf{W} \mathbf{x}_i \ \Vert \ \mathbf{W}, \mathbf{x}_j]) = {\rm LeakyReLU}(\vec{\mathbf{a}}^{\rm T}[\mathbf{W} \mathbf{x}_i \ \Vert \ \mathbf{x}_j])
\end{equation}
where $\mathbf{e}_{ij}$ represents the similarity between node $i$ and $j$; $\Vert$ is the concatenation operation; $\mathbf{W} \in \mathbb{R}^{F' \times F}$ denotes a shared linear transformation that changes the number of input feature from $F$ to $F'$ for each node; and $a : \mathbb{R}^{F’} \times \mathbb{R}^{F'} \rightarrow \mathbb{R}$ represents a shared attention mechanism, which applies the LeakyReLU function \cite{LeaklyReLU2013Maas} to map the concatenated high-dimensional features with a weight vector $\vec{\mathbf{a}} \in \mathbb{R}^{2F'}$ to generate the similarity score.

Step 2: For node $i$, its attention coefficient for each neighbor $j \in \mathcal{N}_i$ is calculated according to Formula \eqref{eq:attention}.

\begin{equation}
	\label{eq:attention}
	\alpha_{ij} = {\rm softmax}_j(\mathbf{e}_{ij}) = \frac{{\rm exp}(\mathbf{e}_{ij})}{\sum_{n \in \mathcal{N}_i} {\rm exp} (\mathbf{e}_{in})}
\end{equation}
where $\alpha_{ij}$ represents the attention coefficient of node $i$ and $j$. The attention coefficient indicates the feature importance of node $j$ to node $i$ and the softmax function \cite{Softmax2022Dubey} is used to normalize it. 


Step 3: For node $i$, $K$ independent attention mechanisms are performed and averaged to generate $\mathbf{x}_i'$ the output of GAT layer according to Formula \eqref{eq:GAT_layer_output}.

\begin{equation}
	\label{eq:GAT_layer_output}
	\mathbf{x}_i' = \sigma (\frac{1}{K} \sum_{k=1}^{K} \sum_{j \in \mathcal{N}_i} \alpha_{ij}^k \mathbf{W}^k \mathbf{x}_j) 
\end{equation}
where $\sigma$ is the activation function (e.g., Sigmoid or ReLU).

As shown in Fig. \ref{fig:overall} (a), by stacking multiple GAT layers, the spatial information of nodes (i.e., zones in the area) can be captured and propagated among their neighbors, e.g., if $M$ GAT layers are used, each node can obtain the knowledge within $M$-hop neighbors. Moreover, to tackle the over-smooth problem caused by the stacked GAT layers, a momentum residual connection \cite{ResNet2016He,APPNP2019Johannes} is introduced. Specifically, as for node $i$, the final output of the graph embedding module $x_i''$ can be calculated according to Formula \eqref{eq:GAT_output}.
\begin{equation}
	\label{eq:GAT_output}
	\mathbf{x_i}'' = \Vert_{m=1}^{M} [(1-\beta) \mathbf{x}_i'^m + \beta \mathbf{x}_i'^{m-1}], \ \mathbf{x}_i'^0 = \mathbf{x}_i
\end{equation}
where $x_i''$ has $MF$ features and $\beta$ is the residual coefficient. Note that due to the data frame constraint of residual connection, the number of features $F$ and $F'$ before and after GAT are set to be the same.


\subsection{Multivariate Decoder Module}

It adopts TPA-LSTM \cite{TPA2019Shih} as the decoder to learn the multivariate time series patterns from the prepared features $\mathbf{x}_{i,t}''$, which represents the features of node $i$ at time $t$. Note that since there is not any node-wise computation in the decoder module, $\mathbf{x}_{i,t}''$ is simplified as $\mathbf{x}_t$ in this section. 

\begin{figure*}
	\centering
	\includegraphics[width=\textwidth]{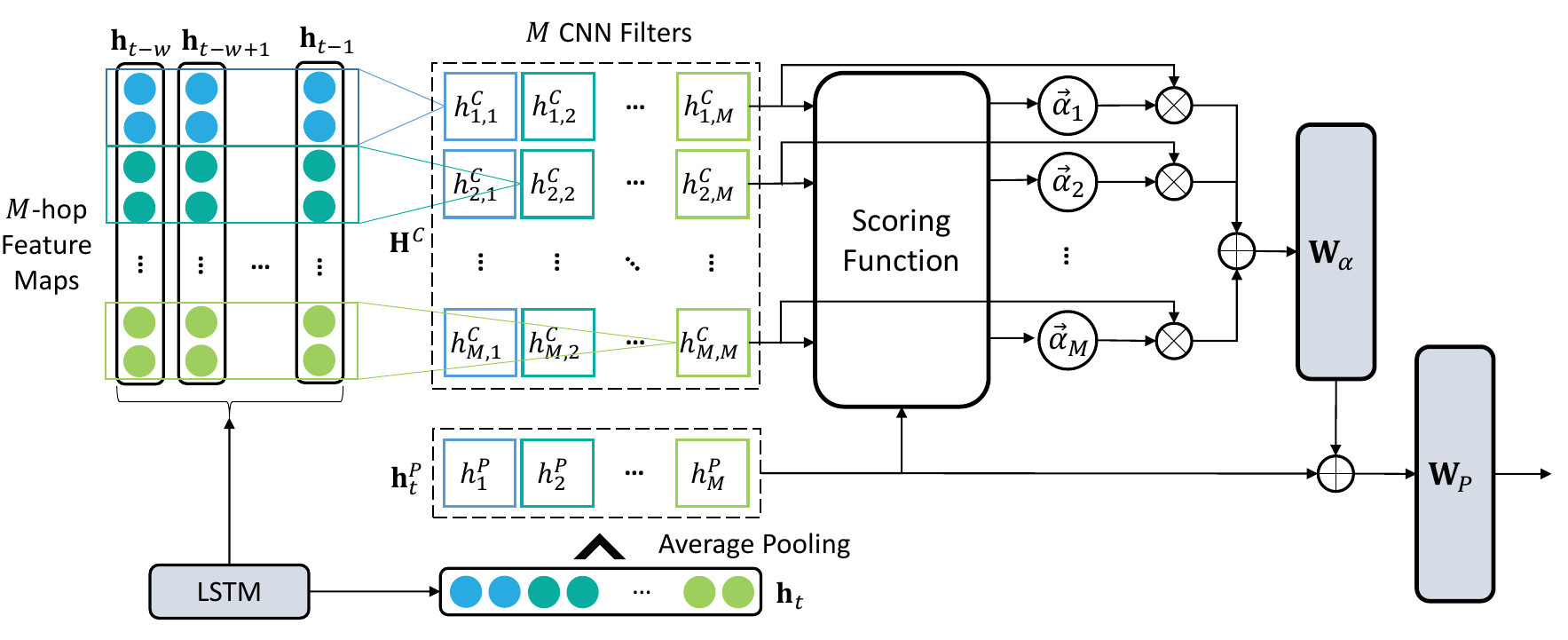}
	\caption{Calculation process of the multivariate temporal decoder module.}
	\label{fig:TPA}
\end{figure*}

Within the decoder, as shown in Fig. \ref{fig:overall} (b), first, a Long Short-Term Memory (LSTM) layer \cite{LSTM1997Hochreiter} is used to extract temporal patterns. Specifically, given an input $\mathbf{x}_t \in \mathbb{R}^{MF}$, the hidden state $\mathbf{h_t}$ and the cell state $\mathbf{c_t}$ at time $t$ can be computed according to Formula \eqref{eq:LSTM}. 

\begin{equation}
	\begin{aligned}
		\label{eq:LSTM}
		\mathbf{u}_t &= \sigma (\mathbf{W}_{uu} \mathbf{x}_t + b_{uu} + \mathbf{W}_{hu} \mathbf{h}_{t-1} + b_{hu})
		\\
		\mathbf{f}_t &= \sigma (\mathbf{W}_{uf} \mathbf{x}_t + b_{uf} + \mathbf{W}_{hf} \mathbf{h}_{t-1} + b_{hf})
		\\
		\mathbf{g}_t &= {\rm tanh} (\mathbf{W}_{ug} \mathbf{x}_t + b_{ug} + \mathbf{W}_{hg} \mathbf{h}_{t-1} + b_{hg})
		\\
		\mathbf{q}_t &= \sigma (\mathbf{W}_{uq} \mathbf{x}_t + b_{uq} + \mathbf{W}_{hq} \mathbf{h}_{t-1} + b_{hq})
		\\
		\mathbf{c}_t &= \mathbf{f}_t \odot \mathbf{c}_{t-1} + \mathbf{u}_t \odot \mathbf{g}_t
		\\
		\mathbf{h}_t &= \mathbf{q}_t \odot {\rm tanh} (\mathbf{c}_t)
	\end{aligned}
\end{equation}
where $\mathbf{h}_{t-1}$ is the hidden state of the LSTM layer at time $t-1$ or the initial hidden state at time $0$; $\mathbf{u}_t$, $\mathbf{f}_t$, $\mathbf{g}_t$, and $\mathbf{q}_t$ are the input, forget, cell and output gates, respectively; $\mathbf{W}$ and $b$ are weights and biases; $\sigma$ is the sigmoid function, and $\odot$ is the Hadamard product. Assuming that $\mathbf{h}_t \in \mathbb{R}^{MF}$, then $\mathbf{u}_t$, $\mathbf{f}_t$, $\mathbf{g}_t$, $\mathbf{q}_t$, and $\mathbf{c}_t \in \mathbb{R}^{MF}$; $\mathbf{W}_{uu}$, $\mathbf{W}_{uf}$, $\mathbf{W}_{ug}$, $\mathbf{W}_{uq} \in \mathbb{R}^{MF \times MF}$; $\mathbf{W}_{hu}$, $\mathbf{W}_{hf}$, $\mathbf{W}_{hg}$ and $\mathbf{W}_{hq} \in \mathbb{R}^{MF \times MF}$.

Second, by applying $M$ filters of CNN $\mathbf{C}_m \in \mathbb{R}^{F \times w}$ with a stride length of $F$ on the hidden states $\{\mathbf{h}_{t-w}, \mathbf{h}_{t-w+1}, ..., \mathbf{h}_{t-1}\}$, and an average pooling $\mathbf{P}_\alpha \in \mathbb{R}^{F \times 1}$ with a stride length of $F$ on $\mathbf{h}_{t}$,  the computational process of TPA can be revised to create attention on both hop-wise and sequence-wise features as illustrated in Fig. \ref{fig:TPA}. Mathematically, $m_{th}$ attention score can be calculated according to Formula \eqref{eq:TPA_attention}.

\begin{equation}
	\label{eq:TPA_attention}
	\vec{\alpha}_m = \sigma ( \mathbf{P}_\alpha^{\rm T} \mathbf{h}_t \odot \Vert_{m=1}^M [\mathbf{C}_m^{\rm T} \mathbf{H}_m]) = \sigma (\mathbf{h}_t^P \odot  \Vert_{m=1}^M [h_{m,m}^C])
\end{equation}
where $\sigma$ is the sigmoid activation function instead of softmax, because it is expected to have more than one variable to support the forecasting; $\mathbf{H}_m$ represents the features with respect to $m_{th}$ hop; $\mathbf{h}_t^P$ and $h_{m,m}^C$ are the feature vector and the $m_{th}$ scalar after the pooling and convolution calculation $\mathbf{P}_\alpha$ and $\mathbf{C}_m$, respectively. In total, there are $M \times M$ attention scores, $\vec{\alpha}_m \in \mathbb{R}^M, m=1,2,...,M$. 

Finally, two linear transformation layers denoted as $\mathbf{W}_P$ and $\mathbf{W}_\alpha \in \mathbb{R}^{M}$, respectively, are introduced to integrate the information for the prediction value $y$ of each node as defined in Formula \eqref{eq:prediction}.

\begin{equation}
	\label{eq:prediction}
	\mathbf{y} = \mathbf{W}_p^{\rm T} (\mathbf{W}_\alpha \mathbf{H}^C \mathbf{\alpha} + \mathbf{h}_t^P)
\end{equation}
where $\mathbf{H}$ is the hidden feature map from time $(t-w)$ to $(t-1)$ and $\mathbf{\alpha} \in \mathbb{R}^{M \times M}$ is their attention scores. 

\subsection{Model Pre-training Module}

\begin{figure}
	\centering
	\includegraphics[width=3.5in]{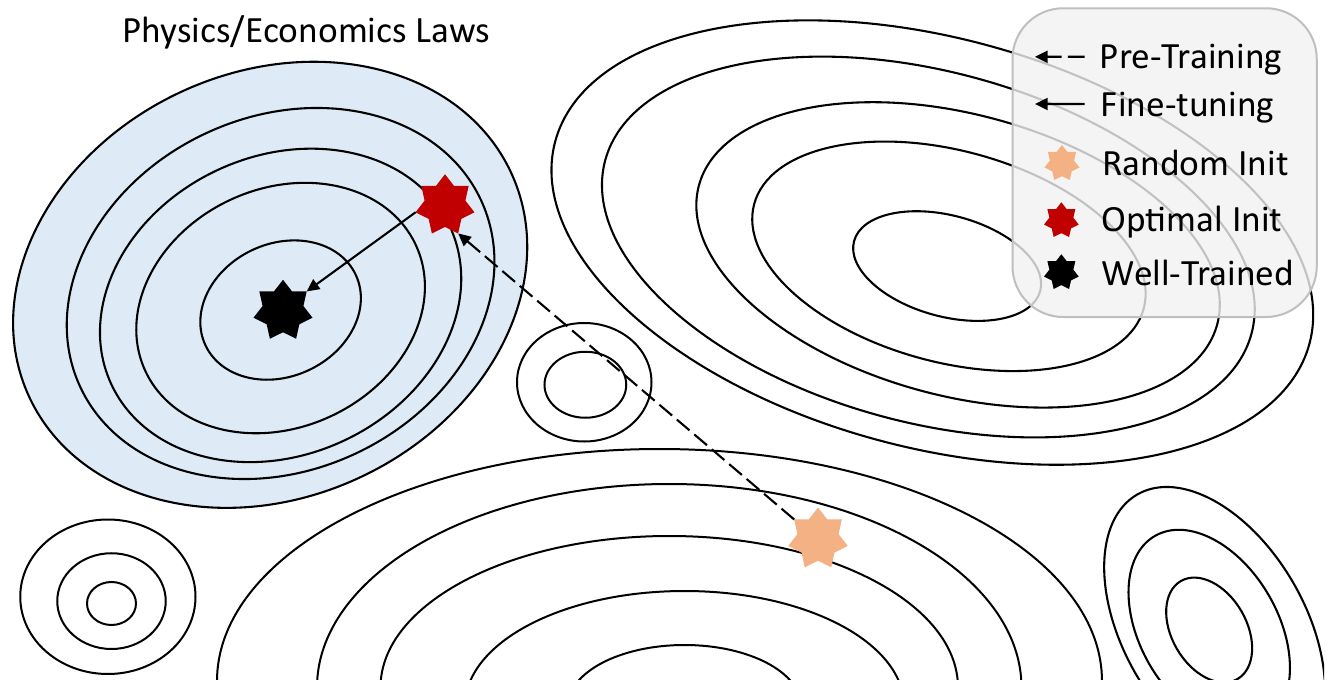}
	\caption{Idea behind the model pre-training framework of Physics-Informed Meta-Learning.}
	\label{fig:gradient}
\end{figure}

As illustrated in Fig. \ref{fig:gradient}, the idea behind the model pre-training module is explained. Assume that a neural network is trained to detect the position of a ball by throwing it randomly. If the observation is not enough or not all situations are experienced, the trained model will be most likely close to the edge of ``Laws'' (i.e., the blue area in the figure) instead of the right central. Such misinterpretations are difficult to be spotted because these local optima usually perform well in most cases. However, misinterpretations in Deep Neural Networks (DNNs) can lead to fatal errors, i.e., in this study, ``a higher price can bring higher charging demands''. To address such misinterpretations, a scalable model pre-training strategy is proposed by combining Physics-informed Neural Networks and Meta-learning.

In general, the key idea is to generate multiple groups of tuning samples based on physical or economic laws, e.g., a consistent price increase will lead to a demand decrease, and then use them to pre-train the model based on meta-learning methods, e.g., First-order Model-agnostic Meta-learning (FOMAML). To address the misinterpretation of the charging price $p$, a set of normally distributed impulses $\Delta p$ is generated, and then corresponding responses $\Delta y$ are calculated based on the physical or economic laws $\mathcal{F}$. For example, according to the recent investigation \cite{EVelasticity2021Bao}, the price elasticity of demand for EV public charging in Beijing (the capital city of China) is approximately $-1.48$, which can be formulated by ($\Delta y / y) / (\Delta p / p) = -1.48$. Given this, the tuning samples of node $i$ can be generated according to Formula \eqref{eq:impulse}.

\begin{equation}
		\label{eq:impulse}
		\Delta y_i = \mathcal{F}(\Delta p_i)=-1.48 (\Delta p_i / p_i) y_i
\end{equation}

Furthermore, the spillover effect between adjacent areas caused by price fluctuations needs to be addressed as well. In general, the fluctuation may impact self and neighboring charging demands differently (i.e., the continuous or intolerable increase of charging prices leads to self-demand decrease but neighboring demand increase). Therefore, assuming that there are $N$ neighbors, the tuning sample of neighbor $j$ can be generated according to Formula \eqref{eq:response}.
\begin{equation}
		\label{eq:response}
		\begin{aligned}
			\Delta y_j = - \frac{1}{N} \Delta y_i
		\end{aligned}
\end{equation}

The two kinds of tuning samples can be merged to form the tuning dataset. Even though such pairs of impulses and responses cannot fully express the complex patterns in the real world, it is sufficient to make the initial parameters inclined to the sensible optimum. Notably, the tuning-sample buffers that reflect knowledge of other restriction laws can be created by repeating the above steps.

As shown in Fig. \ref{fig:overall} (c), a hybrid pre-training strategy based on meta-learning is adopted. First, in each learning iteration, it will gradually reduce the proportion of tuning-samples in each learning iteration by adding observed samples. Second, First-order Model-agnostic Meta-learning (FOMAML) is used to learn knowledge from multiple sample buffers simultaneously. 

To be specific, assuming that there are $S$ buffers, the objective function of FOMAML to find a set of initial parameters minimizing the learner loss is defined by Formula \ref{eq:MAML_obj}.
\begin{equation}
	\label{eq:MAML_obj}
	\min{L(\phi)}=\sum_{s=1}^{S}{l^s({\theta}^s)}
\end{equation}
where $\theta$ is the temporal parameter related to the initial parameter $\phi$; ${\theta}^s$ represents the temporary parameter in buffer $\mathbb{B}_s$; $l^s({\theta}^s)$ is the error calculated by the loss function of the model; and $L(\phi)$ is the total after-training loss of initial parameter $\phi$.

Furthermore, its gradient function can be written as Formula \eqref{eq:MAML_gradient}, where $\nabla_\phi L(\phi)$ is the gradient of $L(\phi)$ with respect to $\phi$; $\nabla_\phi l^s({\theta}^s)$ is the gradient of $l^s({\theta}^s)$ with respect to $\phi$; and $\nabla_\theta l^s({\theta}^s)$ is the gradient of $l^s({\theta}^s)$ with respect to $\theta$.

\begin{equation}
	\label{eq:MAML_gradient}
	\nabla_\phi L(\phi)=\sum_{s=1}^{S}{\nabla_\phi l^s}({\theta}^s)=\sum_{s=1}^{S}{\frac{{\partial l}^s}{{\partial \theta}^s}\frac{{\partial \theta}^s}{\partial \phi}}
\end{equation}

Because the temporary parameter $\theta^s$ is updated by the global initial parameter $\theta$ in buffer $\mathbb{B}_s$, mathematically $\theta^s = \phi - \varepsilon \nabla_\phi l(\phi)$, where $\varepsilon$ is the learning rate, then ${\partial \theta}^s / \partial \phi$ in Formula \eqref{eq:MAML_gradient} can be rewritten to Formula \eqref{eq:approximation}, in which, the second derivative term is omitted to simplify the calculation. The detailed proofs and evaluations of this first-order approximation of meta-learning can be found in \cite{Meta2017Finn}.

\begin{equation}
	\label{eq:approximation}
	\frac{{\partial \theta}^s}{\partial \phi} = 1 - \frac{\varepsilon \nabla_\phi l(\phi)}{\partial \phi \partial \phi} \approx 1
\end{equation}

Therefore, the gradient function of FOMAML can be defined in Formula \eqref{eq:FOMAML}. 

\begin{equation}
	\label{eq:FOMAML}
	\nabla_\phi L(\phi) \approx \sum_{s=1}^{S}{\frac{{\partial l}^s}{{\partial \theta}^s}} =\sum_{s=1}^{S}{\nabla_\theta l^s}({\theta}^s)
\end{equation}

On this basis, the pre-training module updates $\phi_e$ (the initial parameter of the $e_{th}$ epoch) to $\phi_{e+1}$ (the initial parameter of the next epoch). The samples in buffers are divided into two parts, i.e., Support and Query sets, for obtaining intermediate parameters $({\theta}^1,\ldots,\ {\theta}^S)$ and the insightful gradients $\nabla_\phi L(\phi)$, respectively. Finally, $\phi_{e+1}$ is updated according to Formula \eqref{eq:update}, where $\lambda$ is the learning rate.

\begin{equation}
	\label{eq:update}
	\phi_{e+1}=\phi_e- \lambda \frac{\nabla_\phi L(\phi)}{S}
\end{equation}

In general, the pre-training module can be invoked in every epoch until the stop condition is reached, which could be a fixed epoch number or a predefined loss boundary. 

\subsection{Algorithm of the Proposed Approach}

The proposed approach can first run the pre-training module by processing the tuning samples (which are created by the pre-defined physical laws) to prepare optimal initial parameters, and then update the model by using the training samples (a.k.a., the observed samples). Accordingly, the algorithm of the model pre-training module is described in Algorithm \ref{code:model_pretraining}. Note that it is essential to set an appropriate number of pre-training rounds, otherwise, it may lead to model curing, i.e., over-fitting to the tuning samples. 

\begin{algorithm}[h]
	\caption{The model pre-training module}
	\begin{algorithmic}[1]
		\REQUIRE A group of real samples, $S$ laws of physics and economics;
		\STATE Generate $S$ tuning-sample buffers according to the given laws, divide the buffers by time and prepare Support set $\mathcal{R}_1$ and Query set $\mathcal{R}_2$ for each buffer;
		\STATE Initialize the model $\phi$ and learning rate $\lambda$;
		\FOR{each epoch}
		\FOR{each buffer $s$ (in parallel)}
		\STATE Compute $g_1^s = \nabla_\phi l^s(\phi, \mathcal{R}^s_1)$;
		\STATE Update $\phi$ to $\theta^s$ with $g_1^s$;
		\STATE Obtain $g_2^s = \nabla_\theta l^s(\theta^s, \mathcal{R}^s_2)$;
		\STATE Gradually reduce the proportion of tuning samples in each buffer by mixing in real samples;
		\ENDFOR
		\STATE Integrate all the $g_2^s$, where $s=1,2,...,S$
		\STATE Update the global model $\phi$
		\ENDFOR
		\STATE Return a set of pre-trained model parameters $\phi'$
		
	\end{algorithmic}
	\label{code:model_pretraining}
\end{algorithm}

\begin{figure*}
	\centering
	\includegraphics[width=6.8in]{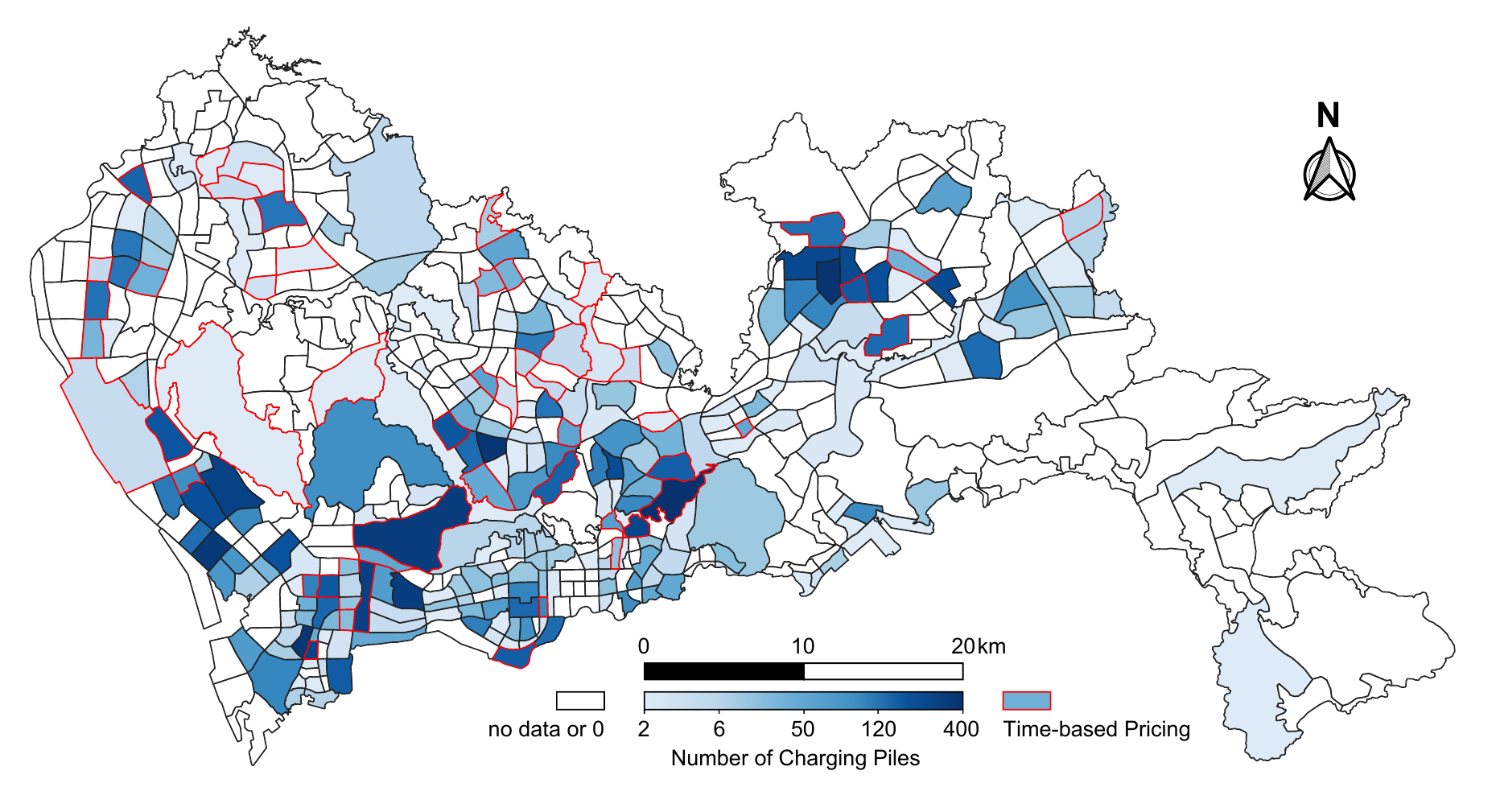}
	\caption{Spatial distribution of public EV charging piles in Shenzhen, China. The studied area contains 247 traffic zones.}
	\label{fig:map}
\end{figure*}

Based on the above-described pre-training process, the overall learning procedure of the proposed approach can be defined and described in Algorithm \ref{code:MetaGTA}. Specifically, according to the hyperparameters, the prediction model is first initialized by the pre-training. After the pre-trained model is created, it is further customized by the observed samples. Once the model is ready, it can be used to predict the EV charging demand of a given zone in the studied area. 

\begin{algorithm}[h]
	\caption{The proposed approach PAG}
	\begin{algorithmic}[1]
		\REQUIRE Hyperparameters (e.g., learning rate, epoch number, loss function, optimizer, and laws of physics and economics), and training samples (a.k.a., observed samples).
		\STATE Initialize the prediction model $\phi$
		\STATE Pre-train the model $\phi$ according to Algorithm \ref{code:model_pretraining} to obtain optimal initial network parameters $\phi'$
		\STATE Fine-tune the pre-trained model $\phi'$ on training set with observed samples
		\STATE Predict the near-future EV charging demand
	\end{algorithmic}
	\label{code:MetaGTA}
\end{algorithm}

In summary, first, the proposed approach utilizes GAT and TPA to construct an integrated network, which can enhance the ability in extracting features from spatiotemporal information. Moreover, it introduces the physics-informed pre-training method based on meta-learning to remedy the impacts of misinterpretations. Such that, the approach can achieve outstanding performance as evaluated in the following section.

\section{Model evaluation} \label{sec:evaluation}

In this section, the proposed approach is evaluated on an EV charging dataset collected in Shenzhen and compared with other representative prediction models. Moreover, ablation experiments are conducted to verify the necessity of each part of the proposed model. Note that the datasets and code used in this paper are shared in Github, and downloadable from the link \footnote{https://github.com/IntelligentSystemsLab/ST-EVCDP}.

\subsection{Evaluation preparation}

\subsubsection{Evaluation data}
The data used in this study is drawn from a publicly available mobile application, which provides the real-time availability of charging piles (i.e., idle or not). Within Shenzhen, China, a total of 18,061 charging piles are covered during the studied period from 19 June to 18 July 2022 (30 days). Besides, the pricing schemes for the studied charging piles are also collected. For analytical purposes, the data is organized as pile occupancy (i.e., demands) and charging price, which are updated every five minutes for the 247 traffic zones introduced by the sixth Residential Travel Survey of Shenzhen \cite{Analysis2019Ma}. As shown in Fig.~\ref{fig:map}, 57 of the zones use a time-based pricing scheme for EV charging, while the remainders use a fixed pricing scheme. From a spatial perspective, all the studied traffic zones are connected with their adjacent neighborhoods to form a graph dataset with 247 nodes and 1006 edges. From a temporal perspective, the evaluation data with a total of 8640 timestamps is divided into training, validation, and test sets with a ratio of 6:2:2 in chronological order, i.e., Day 1-18, Day 19-24, and Day 25-30, respectively. Moreover, each method is configured to run separately to predict the charging demand in four different time intervals, i.e., 15, 30, 45, and 60 minutes. Note that since the temporal resolution of the dataset is 5 minutes, their corresponding $\Delta$ (the fixed time interval for forecasting used in the method) are 3, 6, 9, and 12, respectively.

\begin{table*}[htbp]
	\centering
	\caption{The list of running configurations of compared methods.}
	\begin{threeparttable}
		\begin{tabular}{cccc}
			\toprule
			Model & Setting & Value & Description \\
			\midrule
			\multirow{6}[2]{*}{NNs} & input & (512, 247, 12, 2) & (batch, node, sequence, feature) \\
			& output & (512, 247, 1) & (batch size, node number, 1) \\
			& seed & 2023  & The random seed for parameters initialization \\
			& epoch & 1000  & Early stop if no loss decline for 100 consecutive epochs \\
			& Loss function & MSE   & a widely use loss function for regression tasks \\
			& Optimizer & Adam  & weight decay = 0.00001, Learning rate=0.001 \\
			\midrule
			\multirow{4}[1]{*}{PAG} & $\beta$ & 0.5   & The ratio of residual connection \\
			& $\lambda$ & 0.005  & The learning rate of pre-training \\
			& $\gamma$ & $1-e/1000$   & Proportion of pseudo-samples in the $e$th round of pre-training \\
			& Optimizer & BGD   & The gradient integration method in pre-training \\
			\midrule
			GNNs \tnote{*} & $M$     & 2     & The number of graph propagation layers \\
			GATs \tnote{**} & $K$     & 4     & The number of attention heads \\
			STGCN, DCRNN & -- & -- & Referenced from the original papers \\
			KNN   & (n\_neighbors, leaf\_size) & (5, 30) & It is implemented on Scikit-learn (sklearn) \\
			Lasso & alpha & $1 \times 10^{-9}$ & It is implemented on Scikit-learn (sklearn) \\
			VAR & Number of variables & $247 \times 12 \times 2$ & Namely the number of node, sequence, and feature \\
			\bottomrule
		\end{tabular}%
		\begin{tablenotes}
			\footnotesize
			\item[*] GCN, GAT, GCN-LSTM, AST-GAT, PAG-, and PAG;
			\item[**] GAT, AST-GAT, PAG-, and PAG.
		\end{tablenotes}
	\end{threeparttable}
	\label{tab:configuration}%
\end{table*}%

\subsubsection{The setting of the proposed approach}
PAG is configured as followings: First, the window size $w$ is 12 intervals, which means the model retrospects 60 mins for prediction, as the average travel time for private cars in cities is usually less than one hour \cite{Longitudinal2020Tenkanen}. Second, the two hyper-parameters, i.e., the number of attention heads and GAT layers $K$ and $M$, respectively, are set according to previous studies. Specifically, relevant research on graph propagation \cite{LightGCN2020He} found that $M$ should be much smaller than the radius of the studied graph, and another research on attention mechanism \cite{GAT2018Petar} suggested that $K$ can be equal to a multiple of feature number. Based on pre-experiments, the most appropriate $M$ and $K$ are 2 and 4. Third, for the residual connection between GAT layers, $\beta$ in Formula \eqref{eq:GAT_output} is set as 0.5. Fourth, two tuning sample buffers are generated based on two price elasticities for EV charging, namely 1) $-1.48$, the price elasticity of demand for EV charging in Beijing \cite{EVelasticity2021Bao}; and 2) $-0.228$, the average price elasticity of electricity demand for households across the global covering the period from 1950 to 2014 \cite{PriceElasticity2018Zhu}. To reach the requirement of meta-learning pre-training, the buffers are further divided into two parts by time, i.e., Support set (Day 1-12) and Query set (Day 13-24). Moreover, epoch numbers of the pre-training and fine-tuning processes are set to 200 and 1000, respectively, with an early stopping mechanism: if the validation loss does not decrease for 100 consecutive epochs. Fifth, Mean Square Error (MSE) is used as the loss function. Finally, to improve the learning performance, Adam \cite{Adam2015Diederik} is used with a mini-batch size of 512, a learning rate of 0.001 and a weight decay of 0.00001, respectively.

\subsubsection{Compared methods}
Three statistical models and seven neural networks (NN) are used as baselines. Specifically, the three statistical models include Vector Auto-Regression (VAR) \cite{IRF2013Atsushi}, Least absolute shrinkage and selection operator (Lasso) \cite{Lasso2011Robert} and K-Nearest Neighbor (KNN) \cite{KNN1967Cover}. The NNs can be categorized into four groups, namely 1) a typical NN-based model: Fully Connected Neural Network (FCNN) \cite{FCNN1990Hsu}; 2) an RNN-based model: Long Short-Term Memory (LSTM) \cite{LSTM1997Hochreiter}; 3) two GNN-based models: Graph Convolutional Networks (GCN) \cite{GCN2017Kipf} and Graph Attention Networks (GAT) \cite{GAT2018Petar}; 4) four spatiotemporal models: GCN-LSTM \cite{Parking2019Yang}, STGCN \cite{STGCN2018Yu} and DCRNN \cite{DCRNN2018Li}, and AST-GAT \cite{ASTGAT2022Li}. In addition, the proposed model without pre-training, called PAG-, is used as a compared model as well. To a fair comparison, the specific configurations of the compared models are based on the suggestions of their authors, while the shared training settings are deployed to the same values as mentioned in the previous subsection, i.e., the retrospective window size $w$, maximum epoch number, early stop mechanism, batch size, loss function, and optimizer. In summary, the important running configurations of these models are listed in Table \ref{tab:configuration}.

\subsubsection{Running environment}
Four common metrics for regression are used, i.e., Root Mean Squared Error (RMSE), Mean Absolute Percentage Error (MAPE), Relative Absolute Error (RAE), and Mean Absolute Error (MAE). The result is the average of all nodes (i.e., the 247 studied traffic zones). All the experiments are conducted on a Windows workstation with an NVIDIA Quadro RTX 4000 GPU and an Intel(R) Core(TM) i9-10900K CPU with 64G RAM.

\begin{table*}[htbp]
	\centering
	\caption{Performance comparison in different prediction intervals.}
	\begin{tabular}{l|rrrr|r|rrrr|r}
		\toprule
		Metric ($10^{-2}$) & \multicolumn{5}{c|}{RMSE}             & \multicolumn{5}{c}{MAPE} \\
		Model & \multicolumn{1}{l}{15min} & \multicolumn{1}{l}{30min} & \multicolumn{1}{l}{45min} & \multicolumn{1}{l|}{60min} & \multicolumn{1}{l|}{average} & \multicolumn{1}{l}{15min} & \multicolumn{1}{l}{30min} & \multicolumn{1}{l}{45min} & \multicolumn{1}{l|}{60min} & \multicolumn{1}{l}{average} \\
		\midrule
		VAR & 5.56  & 8.50  & 11.20  & 12.51  & 9.44  & 53.57  & 64.02  & 65.93  & 67.03  & 62.64  \\
		Lasso & 4.35  & 6.33  & 7.97  & 9.34  & 7.00  & 12.28  & 19.43  & 25.64  & 31.56  & 22.23  \\
		KNN   & 4.59  & 6.50  & 7.91  & 9.09  & 7.02  & 11.56  & 16.60  & 20.79  & 24.64  & 18.40  \\
		FCNN  & 3.37  & 5.74  & 7.43  & 8.92  & 6.37  & 9.74  & 15.86  & 22.37  & 28.35  & 19.08  \\
		LSTM  & 3.59  & 5.65  & 8.54  & 9.01  & 6.70  & \textbf{8.91} & 14.94  & 21.00  & 24.15  & 17.25  \\
		GCN   & 4.20  & 6.64  & 8.13  & 8.06  & 6.76  & 40.28  & 46.15  & 49.35  & 52.03  & 46.96  \\
		GAT   & 3.39  & 5.74  & 7.41  & 8.89  & 6.36  & 10.24  & 16.74  & 23.62  & 29.38  & 20.00  \\
		GCN-LSTM & 3.36  & 5.79  & 6.34  & 7.74  & 5.81  & 11.72  & 16.92  & 22.48  & 28.17  & 19.32  \\
		STGCN & 3.55  & 5.58  & 6.91  & 8.74  & 6.20  & 31.41  & 32.58  & 43.07  & 49.62  & 39.17  \\
		DCRNN & 3.72  & 5.40  & 6.79  & 7.82  & 5.93  & 11.32  & 16.04  & 19.94  & \textbf{24.68} & 18.00  \\
		AST-GAT & 3.41  & 5.38  & 6.58  & 7.54  & 5.73  & 9.93  & 18.00  & 17.72  & 27.69  & 18.33  \\
		PAG & \textbf{3.02} & \textbf{5.16}  & \textbf{6.52}  & \textbf{7.21}  & \textbf{5.48}  & 9.34  & \textbf{14.57} & \textbf{17.55} & 26.01  & \textbf{16.87} \\
		\midrule
		Metric ($10^{-2}$) & \multicolumn{5}{c|}{RAE}              & \multicolumn{5}{c}{MAE} \\
		Model & \multicolumn{1}{l}{15min} & \multicolumn{1}{l}{30min} & \multicolumn{1}{l}{45min} & \multicolumn{1}{l|}{60min} & \multicolumn{1}{l|}{average} & \multicolumn{1}{l}{15min} & \multicolumn{1}{l}{30min} & \multicolumn{1}{l}{45min} & \multicolumn{1}{l|}{60min} & \multicolumn{1}{l}{average} \\
		\midrule
		VAR & 45.09  & 56.48  & 61.20  & 62.01  & 56.19  & 8.03  & 10.05  & 10.87  & 11.02  & 9.99  \\
		Lasso & 14.12  & 21.04  & 26.45  & 31.33  & 23.24  & 2.51  & 3.74  & 4.70  & 5.64  & 4.15  \\
		KNN   & 13.97  & 20.42  & 25.43  & 29.70  & 22.38  & 2.49  & 3.63  & 4.52  & 5.42  & 4.02  \\
		FCNN  & 11.06  & 18.64  & 24.16  & 29.34  & 20.80  & 1.97  & 3.32  & 4.30  & 5.29  & 3.72  \\
		LSTM  & 10.66  & 18.14  & 26.78  & 29.07  & 21.16  & 1.90  & 3.23  & 4.76  & 6.36  & 4.06  \\
		GCN   & 35.70  & 41.38  & 45.54  & 45.69  & 42.08  & 6.35  & 7.36  & 8.10  & 9.05  & 7.72  \\
		GAT   & 11.30  & 18.90  & 24.54  & 29.57  & 21.08  & 2.01  & 3.36  & 4.36  & 5.35  & 3.77  \\
		GCN-LSTM & 11.10  & 18.80  & 24.00  & 29.32  & 21.80  & 5.70  & 3.17  & 4.56  & 6.06  & 4.87  \\
		STGCN & 28.06  & 31.87  & 36.81  & 40.45  & 34.30  & 4.99  & 5.67  & 6.54  & 7.34  & 6.14  \\
		DCRNN & 12.80  & 18.97  & 23.81  & 28.94  & 21.13  & 2.28  & 3.37  & 4.23  & 5.13  & 3.75  \\
		AST-GAT & 11.39  & 18.70  & 22.30  & 28.74  & 20.28  & 2.03  & 3.33  & 3.96  & 4.62  & 3.49  \\
		PAG & \textbf{10.58} & \textbf{17.64} & \textbf{22.10} & \textbf{28.21} & \textbf{19.63} & \textbf{1.88} & \textbf{3.14} & \textbf{3.93} & \textbf{4.35} & \textbf{3.33} \\
		\bottomrule
	\end{tabular}%
	\label{tab:performance}%
\end{table*}%

\subsection{Evaluation results and discussions} \label{subse:evaluation}
The performance of evaluated methods is analyzed in three aspects, namely 1) the forecasting error to illustrate how well the model is to predict the future; 2) the ablation results to show the role and necessity of each part of the proposed model; and 3) the model interpretations to demonstrate how plausible the model is in handling price fluctuations.

\subsubsection{Forecasting Error}
As shown in Table \ref{tab:performance}, the evaluation metrics of compared models are summarized. The result shows that the recurrent neural network LSTM outperforms both the three statistical models (i.e., VAR, Lasso, and KNN) and the simple neural network FCNN. While the typical graph learning model GCN has a terrible performance in MAPE, RAE, and MAE, and also GAT, which is more capable in weight allocation than GCN, still performs poorly. By contrast, the combination of RNNs and GNNs (i.e., GCN-LSTM, DCRNN, and AST-GAT) is superior to the recurrent network (i.e., LSTM). It suggests that graph knowledge needs to be coupled with temporal patterns for better performance. Furthermore, the attention-based baseline (i.e., AST-GAT) outperforms the two convolution-based models (i.e., GCN-LSTM and DCRNN), illustrating the superiority of the attention-based network structure in spatio-temporal feature representation.

The result also shows that the proposed approach PAG outperforms other methods in all four metrics with fewer forecasting errors, specifically, 0.0573 in RMSE, 16.87\% in MAPE, 19.63\% in RAE, and 0.0393 in MAE. Compared to VAR, a widely-used statistical model for data analysis, the proposed model has an improvement of 61.3\% on average. Furthermore, PAG performs better than LSTM with 18.21\%, 13.17\%, 11.37\%, and 17.98\% improvements in RMSE, MAPE, RAE, and MAE, respectively. When compared to the state-of-the-art spatio-temporal prediction models (i.e., AST-GAT and DCRNN), the improvement becomes 6.00\% in RMSE, 7.14\% in MAPE, 5.14 \% in RAE, and 8.01\% in MAE. These comparisons illustrate that the graph embedding and multivariate decoding modules can work jointly to achieve optimal performance. 

\subsubsection{Ablation Experiment}
\begin{figure}
	\centering
	\includegraphics[width=3.5in]{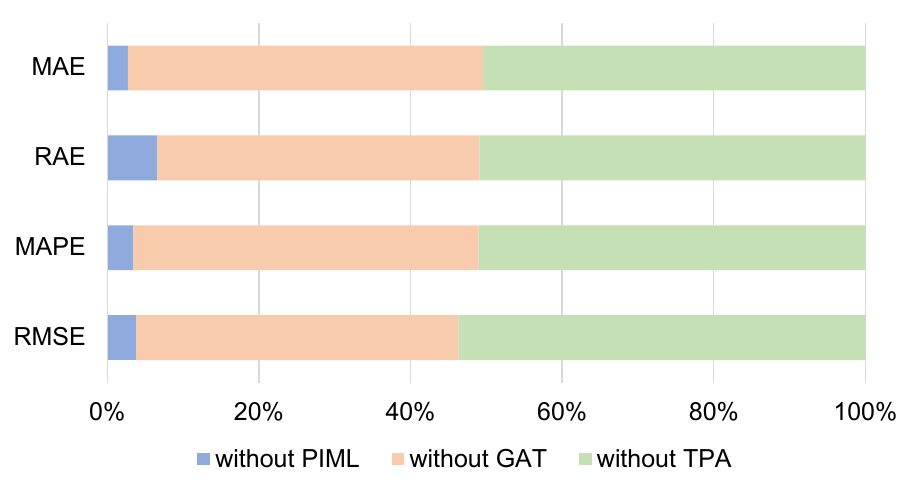}
	\caption{Percent stacked bars of average performance degradation compared to the full model.}
	\label{fig:ablation}
\end{figure}

The ablation experiment is set up by eliminating each module of PAG (i.e., the pre-training step PIML, the graph embedding module GAT, and the multivariate decoder TPA) one by one, and the results are summarized in Table \ref{tab:ablation}. First, it can be concluded from the table that all the proposed modules are necessary for PAG to achieve outstanding performance. Specifically, TPA accounts for the largest contribution, with a decrease in RMSE of 50.84\% compared to the complete model (i..e, noted as PAG), and GAT comes in the second with a 40.27\% decrease in RMSE. To better illustrate the differences, the average degradation of each module is presented in Fig. \ref{fig:ablation}, in which the bars of TPA and GAT make up the majority, while the ones of PIML are marginal. 

To sum up, first, it is efficient and effective by combining GAT and TPA. Second, the underlying spatiotemporal features cannot be extracted by graph or temporal attention alone, as the models without GAT or TPA give similar results to FCNN as shown in Table \ref{tab:performance}. Finally, besides the average improvement achieved by PIML of about 4.18\%, the main contribution of PIML is to reduce misinterpretation, which will be evaluated and discussed in the following subsection.

\begin{table}[htbp]
	\centering
	\caption{Results of ablation experiment.}
	\begin{tabular}{cccccc}
		\toprule
	    Interval & Model & RMSE  & MAPE  & RAE   & MAE \\
		\midrule
		\multirow{4}[2]{*}{15min} & PAG & 3.02  & 9.34  & 10.58  & 1.88  \\
		& without Meta & 3.11  & 9.98  & 11.12  & 1.95  \\
		& without GAT & 4.44  & 16.94  & 15.56  & 3.19  \\
		& without TPA & 5.08  & 17.62  & 16.45  & 3.31  \\
		\midrule
		\multirow{4}[2]{*}{35min} & PAG & 5.16  & 14.94  & 18.14  & 3.23  \\
		& without Meta & 5.33  & 15.28  & 18.76  & 3.34  \\
		& without GAT & 7.61  & 20.14  & 23.19  & 6.33  \\
		& without TPA & 8.14  & 21.24  & 24.77  & 6.79  \\
		\midrule
		\multirow{4}[2]{*}{45min} & PAG & 6.52  & 17.55  & 22.10  & 3.93  \\
		& without Meta & 6.87  & 18.07  & 23.66  & 4.30  \\
		& without GAT & 8.68  & 23.55  & 27.94  & 7.49  \\
		& without TPA & 8.73  & 23.88  & 28.72  & 7.59  \\
		\midrule
		\multirow{4}[2]{*}{60min} & PAG & 7.21  & 26.01  & 28.21  & 4.35  \\
		& without Meta & 7.41  & 26.23  & 29.23  & 4.53  \\
		& without GAT & 9.61  & 31.41  & 34.58  & 8.76  \\
		& without TPA & 10.36  & 32.47  & 35.66  & 8.93  \\
		\bottomrule
	\end{tabular}%
	\label{tab:ablation}%
\end{table}%

\subsubsection{Interpretation}
In this part, whether PAG has a proper interpretation between EV charging demands and prices is discussed. 57 of the studied zones are with dynamic pricing, i.e., the areas with red boxes shown in Fig. \ref{fig:map}. Moreover, the 30-mins prediction model is used for this testing, because the average commuting time in Shenzhen is about 36 minutes according to the 2022 China Major Cities Commuting Monitoring Report \cite{ACMR2022}. A typical model (i.e., GCN-LSTM) and a recently developed model (i.e., AST-GAT) for spatiotemporal prediction are used for comparison.
 
\begin{figure}
	\centering
	\includegraphics[width=3.5in]{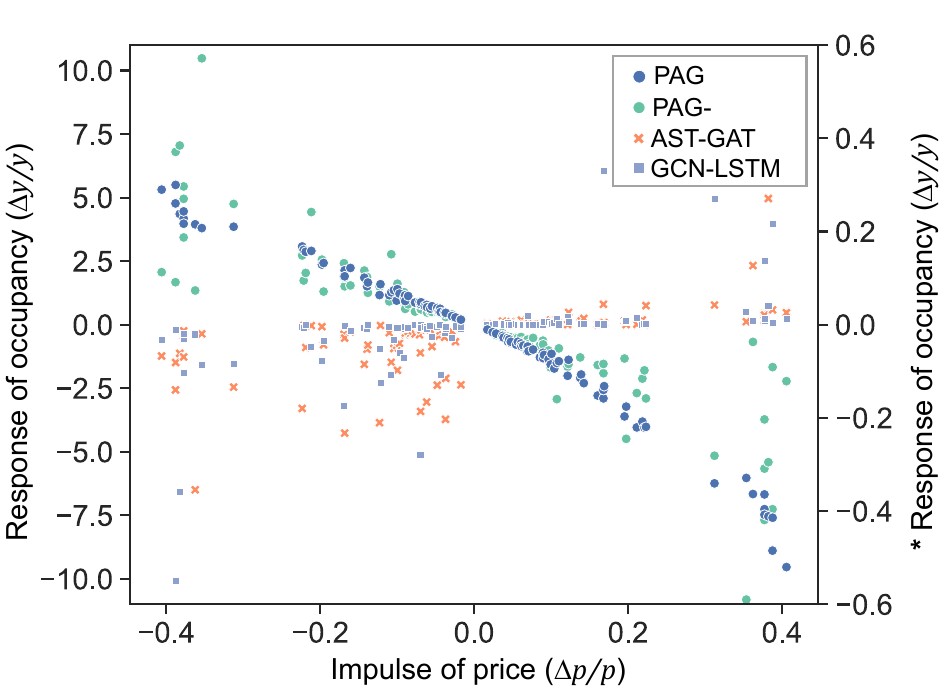}
	\caption{Responses of EV charging occupancy to the std impulses of prices.}
	\label{fig:std_impulse}
\end{figure}

Two types of price impulses are added for the prediction, namely 1) pairs of negative and positive standard deviation of local charging prices, i.e., $\Delta p = {\rm std} (p)$; 2) fixed percentiles of local charging prices, i.e., $\{\pm10\%, \pm 20\%, \pm 30\%\}$. First, Fig. \ref{fig:std_impulse} shows the responses of EV charging occupancy to the standard-deviation impulses of electricity prices. It can be seen that the demand responses of GCN-LSTM and AST-GAT have the same trend encoded in the price impulses, supporting the argument that the existing models over-fit sample distribution and make misinterpretations in the relationship between EV charging demand and price. 

\begin{figure*}
	\centering
	\includegraphics[width=7in]{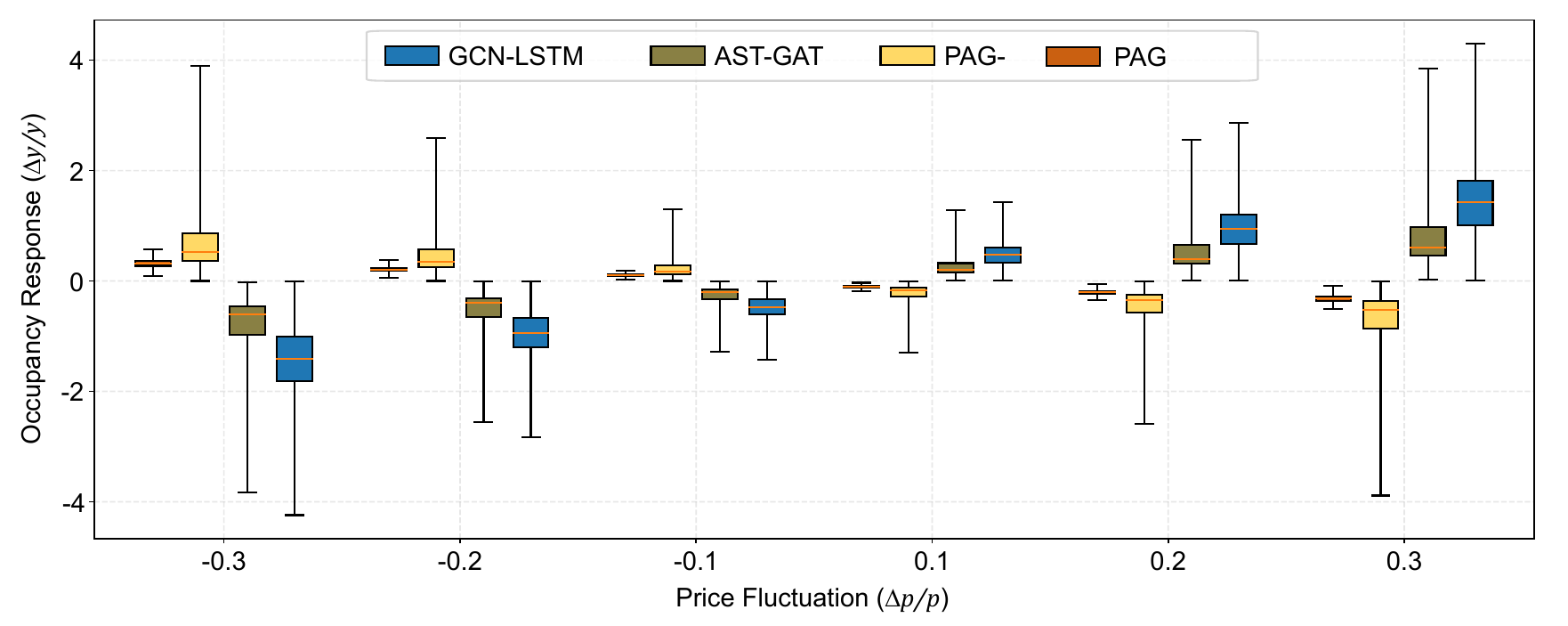}
	\caption{Responses of EV charging occupancy to the percentile impulses of prices.}
	\label{fig:per_impulse}
\end{figure*}

Furthermore, although the attention-based model PAG- is aware of the reverse impact, it reacts erratically up to 1000\% and down to $-$1000\%, approximately, while the price fluctuation ranges between $\pm$40\%. Such a drastic reaction is clearly misbehaved, because energy consumption is usually inelastic to price according to the relevant literature, e.g., natural gas for households\cite{Gas2016Paul}, gasoline for vehicle miles traveled \cite{Gasoline2017Rivers}, and residential electricity demand \cite{PriceElasticity2018Zhu}. Similar phenomena can be seen in Fig. \ref{fig:per_impulse}, where each box shows the distribution of demand responses to the percentile impulses for the 57 traffic zones with time-varying pricing schemes. In contrast, the pre-trained model PAG can give reasonable responses as shown in Fig. \ref{fig:std_impulse}, where PAG has the most concentrated distribution of responses with absolute values ranging from 0 to 0.6, and also in Fig. \ref{fig:per_impulse}, where PAG has minimal average response about 75\% of the corresponding impulse strength. All these results indicate that PAG can interpret an inelastic demand for EV charging with respect to price fluctuations in the studied city (i.e., Shenzhen). 

\begin{figure}
	\centering
	\includegraphics[width=3.5in]{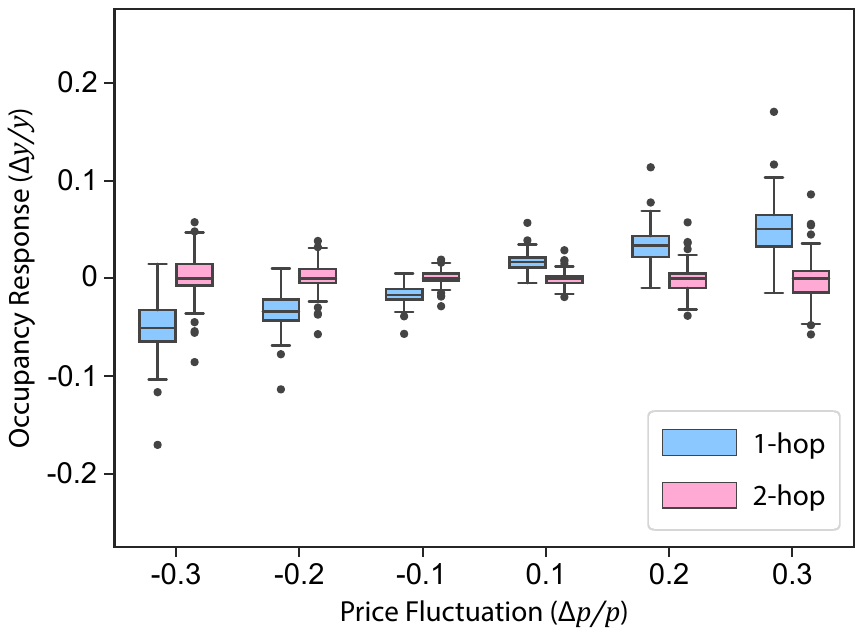}
	\caption{The proposed model's responses in neighboring areas to price fluctuations.}
	\label{fig:spatial}
\end{figure}

\begin{figure}
	\centering
	\includegraphics[width=3.5in]{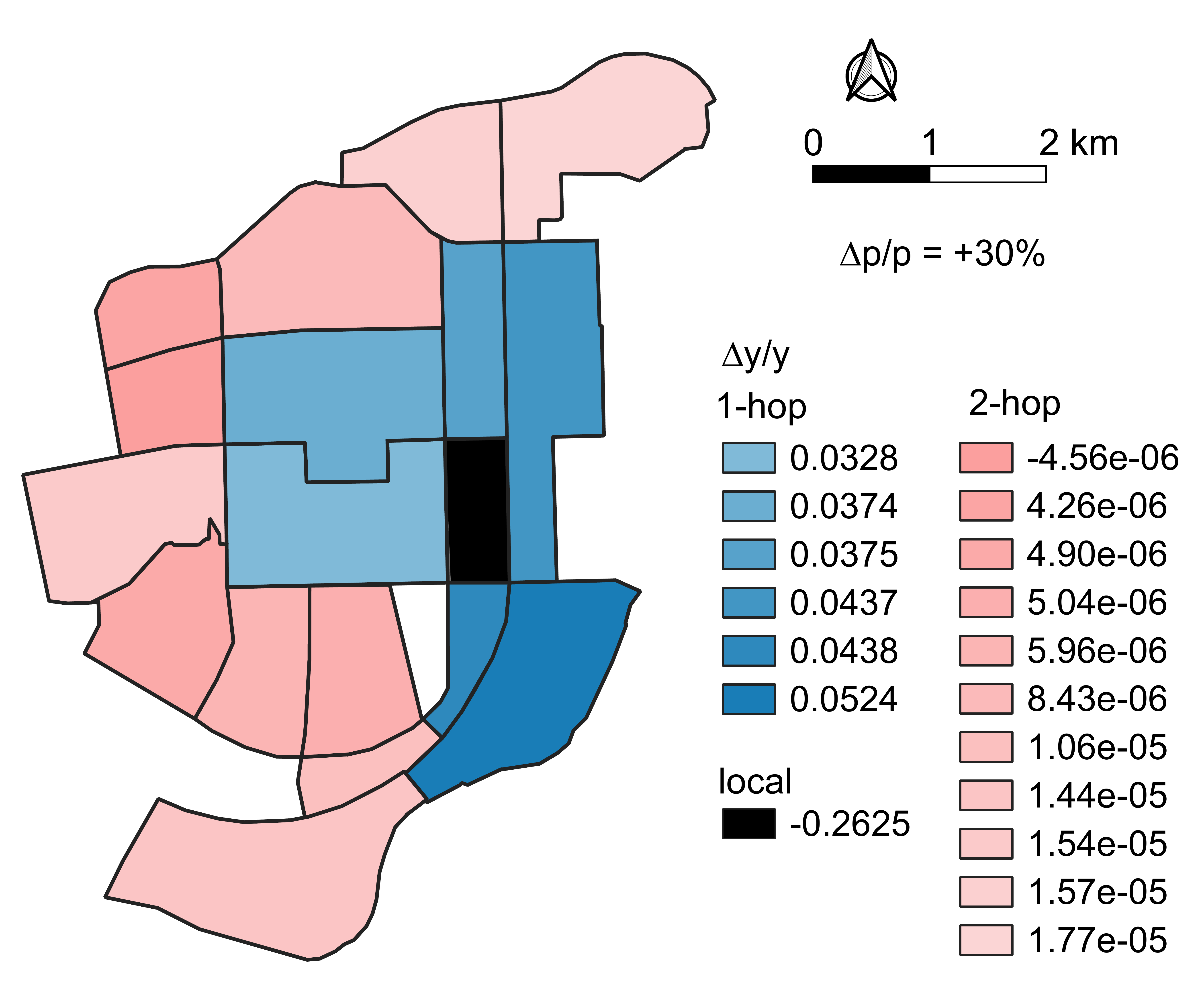}
	\caption{Example of neighboring responses in the central business district of the studied city, Shenzhen.}
	\label{fig:neighboring}
\end{figure}

Third, the price-fluctuation impacts on EV charging demand in 1- and 2-hop neighbors is presented in Fig. \ref{fig:spatial}, where the demand changes in 1-hop adjacent zones have the same sign with the local price fluctuations. It means an increase in charging prices in one area will raise EV charging demand in its neighborhoods, and vice versa. While in its 2-hop neighbors, the direction of the impacts is less certain. Taking a traffic zone in the central business district of Shenzhen as an example, the local and neighborhood responses to a 30\% price change are illustrated in Fig. \ref{fig:neighboring}. It can be seen that the corresponding increases in 1-hop neighborhood demand can offset the reductions in local demand with a ratio of more than 90\%, indicating that surrounding areas can be used as alternatives for related demands. In contrast, it shows that 2-hop neighbors may not be substitutes to accommodate the shifting demands because their responses to price fluctuations are marginal. Given the above results, it can be concluded that the proposed model can obtain a correct understanding of spillover effects to reduce the misinterpretations that are commonly existed in current methods.

In summary, the proposed approach can significantly reduce the prediction errors with the highest scores in all four metrics, i.e., RMSE 0.0548, MAPE 16.87\%, RAE 19.63\%, and MAE 0.0333. In particular, the combination of GAT and TPA for graph embedding and multivariate spatiotemporal decoding is shown to be effective according to the ablation experiments. Moreover, although it is shown that the gains from the pre-training module are marginal, it can prevent misinterpretations to handle the impacts of price fluctuations on charging demands. 

\section{Conclusions and future works} \label{sec:conclusion}
EV charging demand prediction plays an important role in improving the "smartness" of urban power grids and intelligent transportation systems, e.g., directing EV drivers to available parking spots with user experience improved and adjusting charging prices to redirect the charging demand to the surrounding areas. This paper proposes a novel approach, named PAG, to train a prediction model with misinterpretations addressed and performance improved. Specifically, PAG combines graph and temporal attention mechanisms for graph embedding and multivariate time-series decoding and also introduces a model pre-training model based on physics-informed meta-learning. As shown by the evaluation results, PAG can not only outperform the state-of-the-art methods by approximately 6.57\% on average in forecasting errors but also obtain a correct understanding of spillover effects to reduce the misinterpretations between charging demands and prices that are commonly existed in current methods.

Although PAG can achieve state-of-the-art performance, it still has certain limitations and can be improved in future work. First, the focus of this work remains on the relationship between EV charging demand and price. However, there are still many misinterpretations about the current deep learning-based models to assist intelligent transportation systems, especially in spatiotemporal scenarios. Further investigations and corresponding optimizations on the stacked model are needed. Moreover, the model pre-training needs to be carefully monitored to obtain the correct knowledge, and the introduction of an automated critique module trained by Reinforcement Learning may be one way to address this issue. Finally, the approach can be enhanced to quantify the spillover effects. Such that regulators and managers can better detect impacts in related policies to optimize the running of intelligent transportation systems.



\ifCLASSOPTIONcaptionsoff
  \newpage
\fi



%

\bibliographystyle{IEEEtran}
\bibliography{IEEEabrv}



%

\begin{IEEEbiography}[{\includegraphics[width=1in,height=1.25in,clip,keepaspectratio]{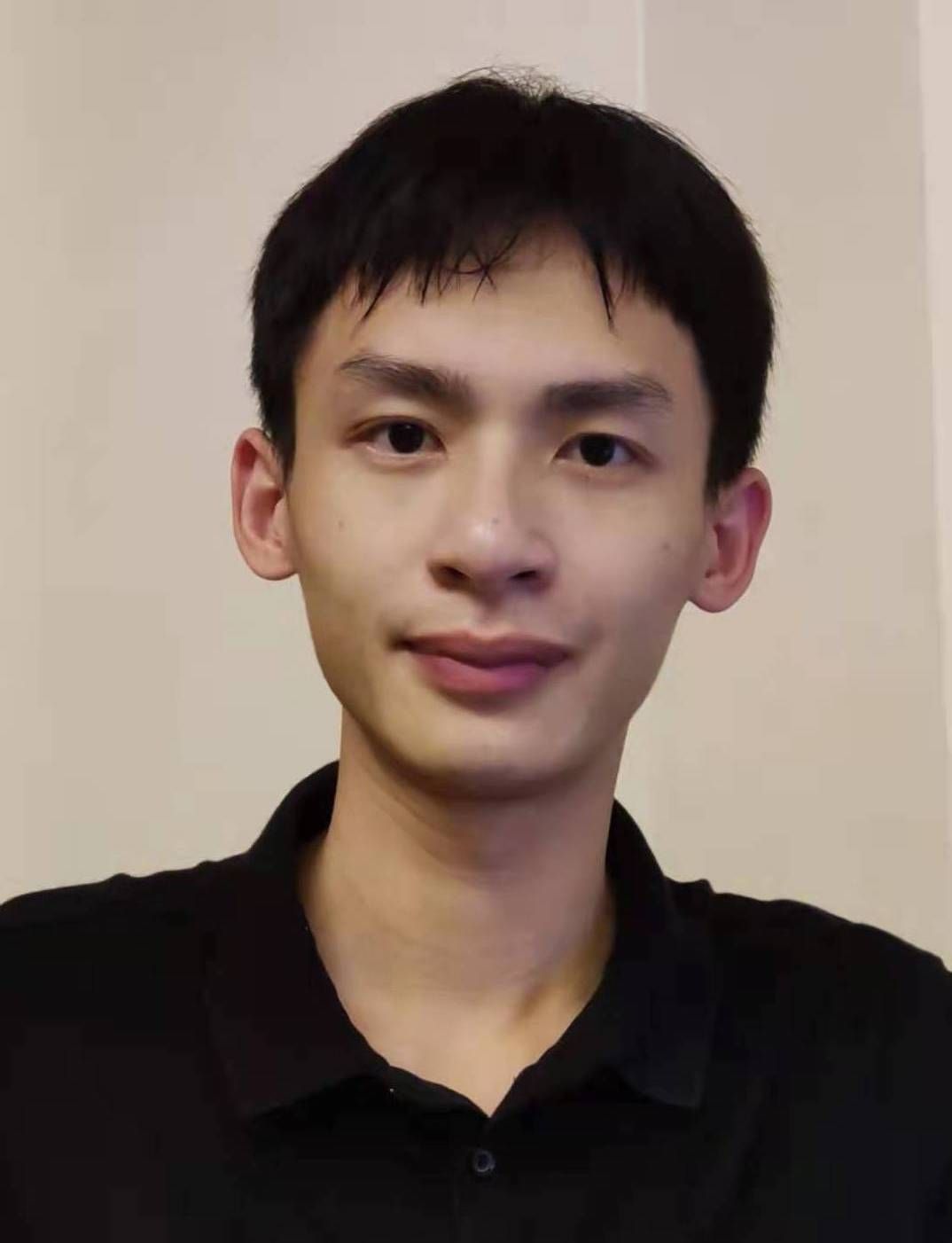}}]{Haohao Qu} is currently a PhD student of the Department of Computing (COMP), The Hong Kong Polytechnic University. He received the B.E. and M.E. degree from the School of Intelligent Systems Engineering, Sun Yat-Sen University, People's Republic of China, in 2019 and 2022, respectively. His research interests include Meta-learning, Graph Neural Networks, and their applications in Intelligent Transportation Systems and Recommendation Systems. He has authored and co-authored innovative works in top-tier journals (e.g., TITS) and international conferences (e.g., KSEM, UIC, and ICTAI). More information about him can be found at https://quhaoh233.github.io/page/.
\end{IEEEbiography}

\begin{IEEEbiography}[{\includegraphics[width=1in,height=1.25in,clip,keepaspectratio]{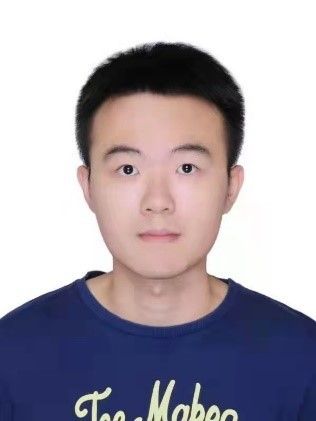}}]{Haoxuan Kuang} received his B.E. degree in School of Intelligent Systems Engineering, Sun Yat-sen University, China, in 2022, where he is currently working toward the M.E. degree in Transportation engineering. His research interests include artificial intelligence, intelligent transportation systems and traffic status prediction.
\end{IEEEbiography}

\begin{IEEEbiography}[{\includegraphics[width=1in,height=1.25in,clip,keepaspectratio]{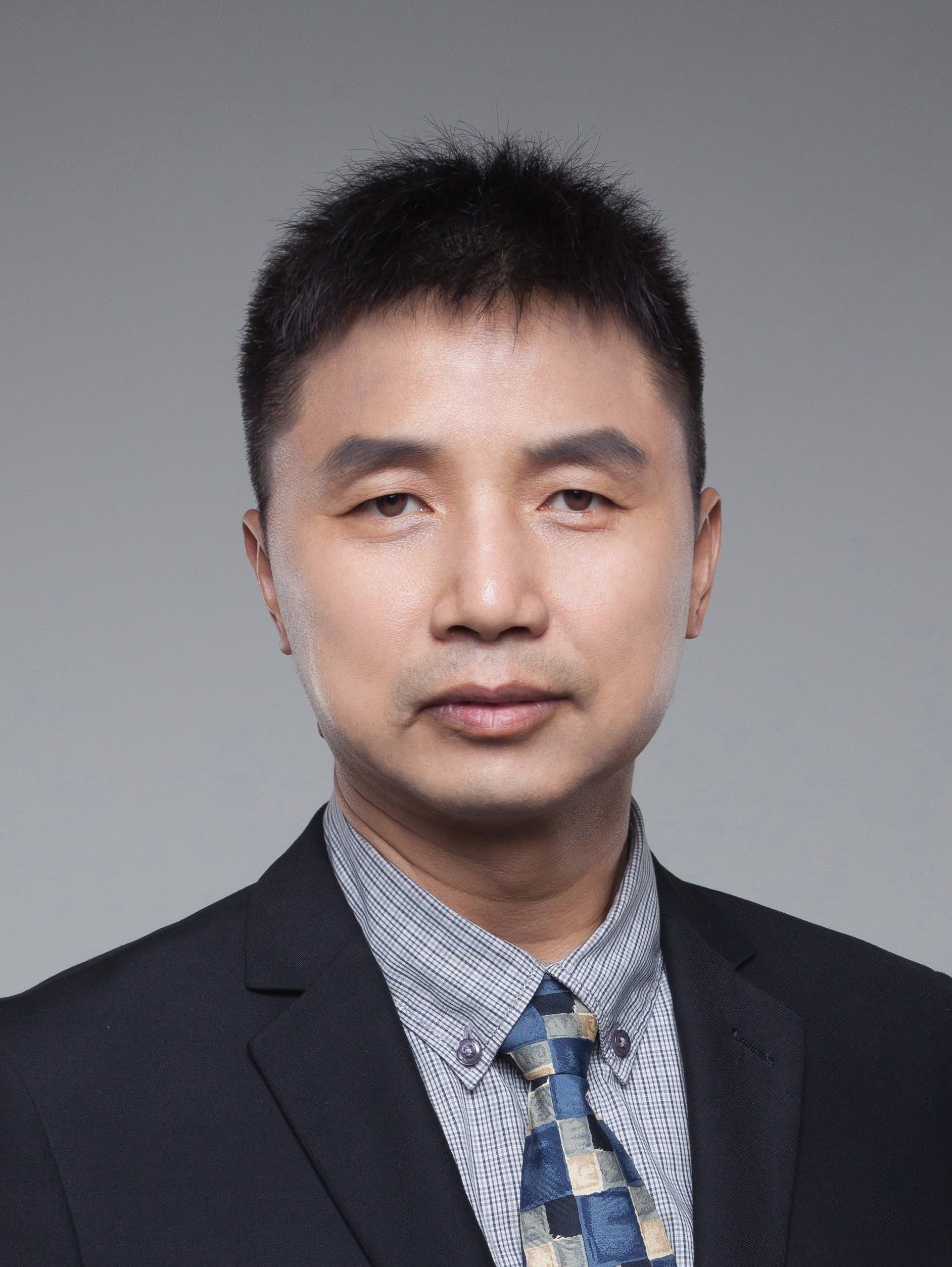}}]{Dr. Jun Li} an Associate Professor at School of Intelligent Systems Engineering, Sun Yat-Sen University. He received his Ph.D. in Civil Engineering from Nagoya University, his master's degree from the Asian Institute of Technology, and his BSc from Tsinghua University. His research interests include travel behavior analysis, transportation economic analysis, and system optimization.
\end{IEEEbiography}

\begin{IEEEbiography}[{\includegraphics[width=1in,height=1.25in,clip,keepaspectratio]{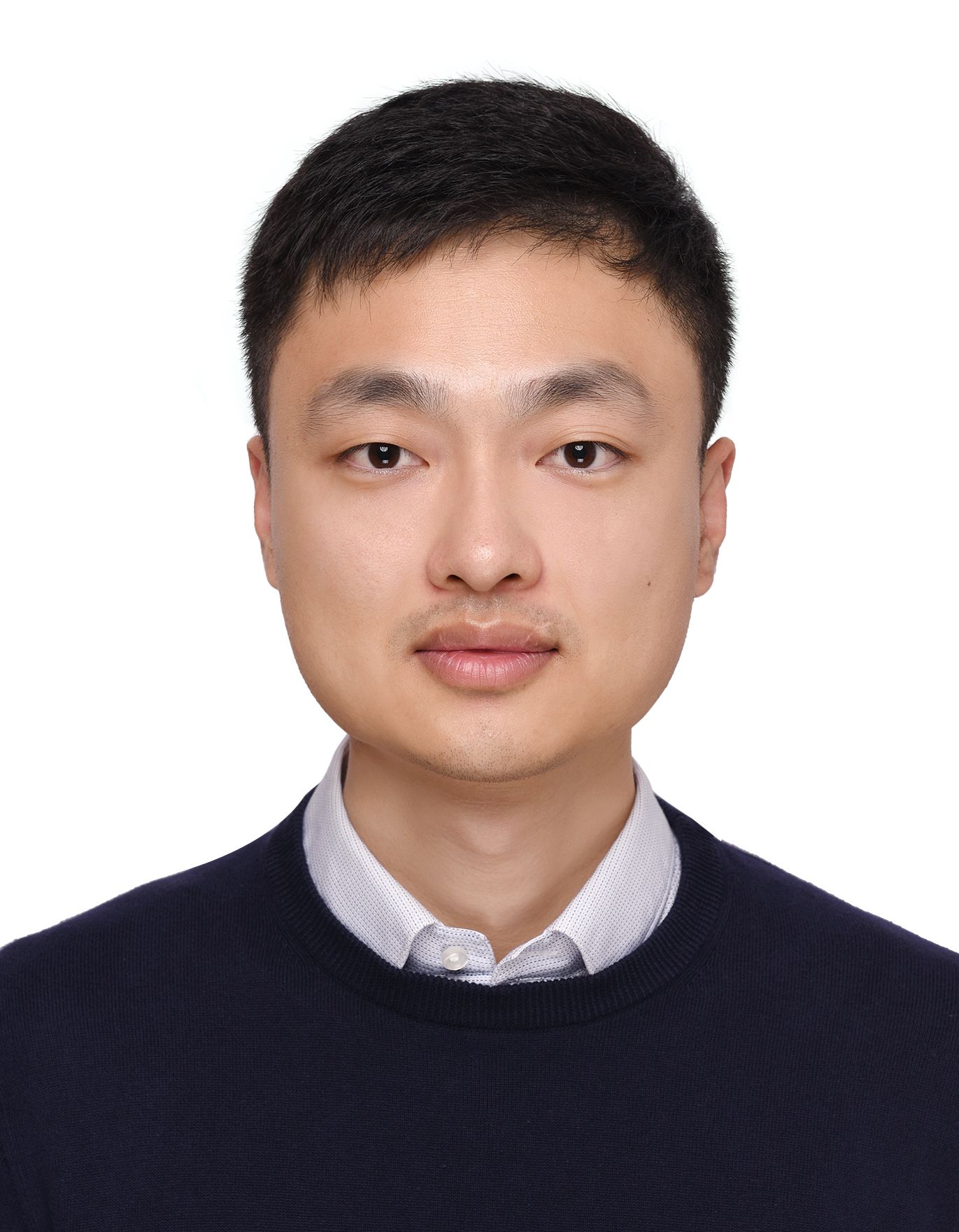}}]{Dr. Linlin You} is an Associate Professor at the School of Intelligent Systems Engineering, Sun Yat-sen University, and a Research Affiliate at the Intelligent Transportation System Lab, Massachusetts Institute of Technology. He was a Senior Postdoc at the Singapore-MIT Alliance for Research and Technology, and a Research Fellow at the Singapore University of Technology and Design. He received his Ph.D. in Computer Science from the University of Pavia, his dual master's degrees with honor in Software Engineering and Computer Science from the Harbin Institute of Technology and the University of Pavia, and his BSc in Software Engineering from the Harbin Institute of Technology. He authored or co-authored more than 40 journal and conference papers in the research fields of Internet of Things, Smart Cities, Autonomous Transportation System, Multi-source Data Fusion, and Federated Learning.
\end{IEEEbiography}






\end{document}